\documentclass[conference]{IEEEtran}
\IEEEoverridecommandlockouts
\usepackage{hyperref}
\usepackage{cite}
\usepackage{amsmath,amssymb,amsfonts}
\usepackage{algorithmic}
\usepackage{graphicx}
\usepackage{textcomp}
\usepackage{xcolor}
\usepackage{multirow}
\def\BibTeX{{\rm B\kern-.05em{\sc i\kern-.025em b}\kern-.08em
    T\kern-.1667em\lower.7ex\hbox{E}\kern-.125emX}}

\begin{document}

\title{Multi-Grained Feature Pruning for Video-Based Human Pose Estimation}
%Multi-Scale Feature Selection for Video-Based Human Pose Estimation

%{\footnotesize \textsuperscript{*}Note: Sub-titles are not captured for https://ieeexplore.ieee.org  and should not be used}\thanks{Identify applicable funding agency here. If none, delete this.}

\author{\IEEEauthorblockN{Zhigang Wang$^{1}$, Shaojing Fan$^{2}$, Zhenguang Liu$^{3, 4 }$, Zheqi Wu$^{1}$, Sifan Wu$^{5}$, Yingying Jiao$^{6 *}$\thanks{This work is supported by the National Natural Science Foundation of China (No. 62372402), and the Key R\&D Program of Zhejiang Province (No. 2023C01217).   * Corresponding author.}}\\
\IEEEauthorblockA{
$^{1}$ College of Computer Science and Technology, Zhejiang Gongshang University, Hangzhou, China\\
$^{2}$ School of Computing, National University of Singapore, Singapore\\
$^{3}$ The State Key Laboratory of Blockchain and Data Security, Zhejiang University, Hangzhou, China\\
$^{4}$ Hangzhou High-Tech Zone (Binjiang) Institute of Blockchain and Data Security, Hangzhou, China\\
$^{5}$ College of Computer Science and Technology, Jilin University, Changchun, China\\
$^{6}$ College of Computer Science and Technology, Zhejiang University of Technology, Hangzhou, China\\
\{wangzhigang2024, fanshaojing, liuzhenguang2008, chasewoo17, wusifan2021\}@gmail.com, jiaoyy21@mails.jlu.edu.cn}}

\maketitle

% To tackle these problems, we propose a novel multi-grained feature prune framework that integrate feature token prune strategy and multi-frained representations learning into a pose estimation model. 
% Video-based human pose estimation has long been a fundamental and compelling challenge in computer vision.
\begin{abstract}

Human pose estimation, with its broad applications in action recognition and motion capture, has experienced significant advancements. However, current Transformer-based methods for video pose estimation often face challenges in managing redundant temporal information and achieving fine-grained perception because they only focus on processing low-resolution features. To address these challenges, we propose a novel multi-scale resolution framework that encodes spatio-temporal representations at varying granularities and executes fine-grained perception compensation. Furthermore, we employ a density peaks clustering method to dynamically identify and prioritize tokens that offer important semantic information. This strategy effectively prunes redundant feature tokens, especially those arising from multi-frame features, thereby optimizing computational efficiency without sacrificing semantic richness. Empirically, it sets new benchmarks for both performance and efficiency on three large-scale datasets. Our method achieves a 93.8\% improvement in inference speed compared to the baseline, while also enhancing pose estimation accuracy, reaching 87.4 mAP on the PoseTrack2017 dataset.

% Fueled by the success of Transformers, numerous video-based human pose estimation methods have utilized them to model temporal contexts and achieve impressive results. However, one aspect that has been overlooked so far is that these methods directly aggregate all temporal feature tokens from neighboring frames, resulting in learning substantial redundant information and consuming significant computational resources. Furthermore, While modelling high-resolution representations is crucial for accurate pose estimation, current Transformer-based methods still maintain compressed low-resolution representations, leading to a deficit of fine-grained perception. To tackle these problems, we propose a novel multi-scale resolution framework that encodes spatio-temporal representations at different granularities and performs multi-grained fusion. We further introduce density peaks clustering to dynamically select representative tokens with high semantic diversity, performing feature token pruning while eliminating redundancy in multi-frame features and improving model efficiency. Empirically, our approach establishes a new benchmark for performance and efficiency on three large-scale datasets.      

\end{abstract}


\section{Introduction}
Human pose estimation, as a fundamental and compelling challenge in the realm of computer vision~\cite{wang2022contextual, geng2023human, wu2024pose, liu2022copy, su2021motion}, aims to precisely locate the anatomical human joints. Now, accurate pose estimation plays a pivotal role in human-computer interaction scenarios, enabling machines to gain a more intelligent grasp of human behaviors and body positions. Accordingly, human pose estimation has garnered substantial attention and spans numerous real-world applications, ranging from \textit{action recognition} to \textit{augmented reality}~\cite{schmidtke2021unsupervised, tse2022collaborative,yang2023action,shuai2023locate}.

%In this paper, we strive to explore the boundaries of performance and efficiency in addressing the challenges of human pose estimation in video sequences. 
A multitude of methods focus on the field of pose estimation in static images, progressing from early methods involving tree-based models~\cite{zhang2009efficient,sapp2010cascaded} to modern approaches leveraging convolutional neural networks~\cite{newell2016hourglass, sun2019hrnet}. However, these image-based methods often overlook the temporal correlations between sequential video frames, which can result in diminished performance when directly applied to the task of video pose estimation. To tackle this issue, one line of work~\cite{ilg2017flownet, doering2018jointflow, bertasius2019posewarper, wu2024joint, chen2025causal} attempts to aggregate temporal motion context, utilizing optical flow or pose residual modelling. Despite their impressive success in videos, these methods easily hit a performance ceiling and deliver suboptimal outcomes, especially when processing low-quality videos characterized by occluded poses and blurred frames. 
% in challenging video scenarios such as severe pose occlusion and fast motion.   
% Despite their remarkable success in video analysis, these methods encounter a performance ceiling and often yield subpar outcomes, particularly when dealing with low-quality videos characterized by obscured poses and blurry frames.

\begin{figure*}[t]
\centering
\includegraphics[width=.9\linewidth]{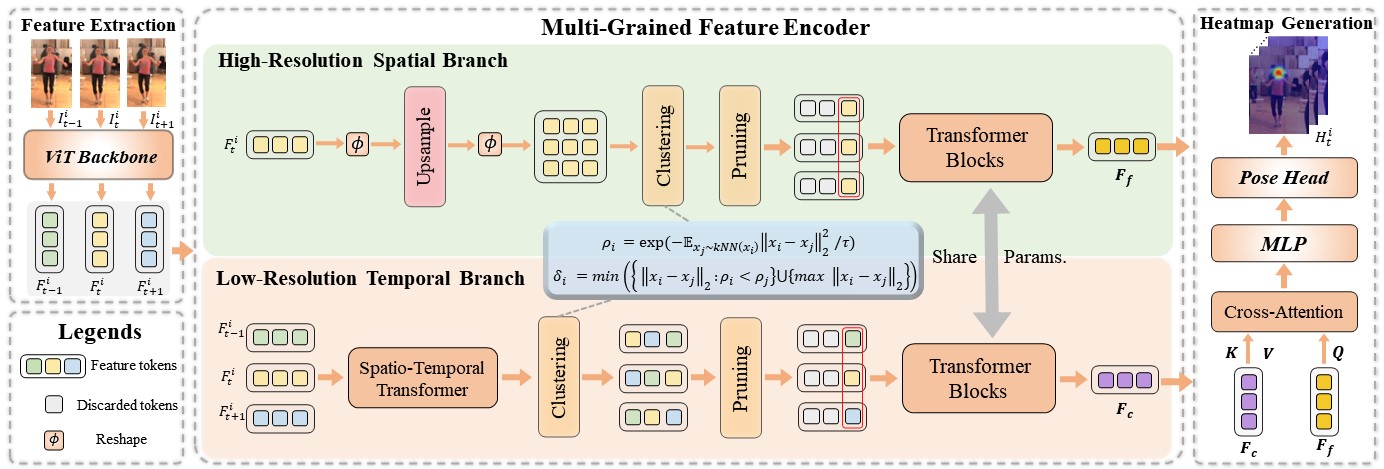}
\caption{The overall pipeline of our FTP-Pose framework. Given an input sequence $\{I_{t-1}^{i}, I_{t}^{i}, I_{t+1}^{i}\}$, our goal is to generate the pose heatmap $\mathbf{H}_{t}^{i}$ of the key frame $I_{t}^{i}$. Initially, we extract the feature tokens $\{\boldsymbol{F}_{t-1}^{i}, \boldsymbol{F}_{t}^{i}, \boldsymbol{F}_{t+1}^{i}\}$ via a ViT backbone. We then feed these features into the Multi-Grained Feature Encoder (MGFE) to manage fine-grained spatial dependencies and capture high-dimensional temporal contexts. Subsequently, the output features derived from MGFE are combined through a cross-attention layer. Finally, these combined features are processed through a specific pose head to estimate the pose heatmaps $\mathbf{H}_{t}^{i}$.}
\label{fig:framework}
\end{figure*}

Recently, Transformer architectures~\cite{vaswani2017transformers, dosovitskiy2020imagevit} have excelled in the field of visual perception tasks, yet their potential in video-based human pose estimation remains to be fully explored. Through careful empirical analysis of existing Transformer-based methods~\cite{he2024dsta, feng2023diffpose, jiao2025spatiotemporalstdpose}, we noticed that they tend to overlook two important aspects: (1) Since video data includes an additional temporal dimension compared to image data, current methods attempt to extract useful temporal information from neighboring frames to aid pose estimation in target frames. However, most existing methods directly concatenate multi-frame feature tokens and compute attention across all tokens, resulting in substantial computational cost and extensive redundant information from highly similar consecutive video frames. (2) Current Transformer-based models~\cite{jin2022otpose, feng2023diffpose, jiao2025optimizingvremd} compress images into embeddings that are 1/16 the size of the target pose heatmap before feeding them into the model, which may lead to a lack of fine-grained visual cues derived from high-resolution feature learning.

% A fact has been demonstrated is that the model architectures, which maintain high-resolution feature lenrning branch, can retain fine-grained visual details.
Motivated by these limitations, we introduce a novel framework, termed \textbf{FTP-Pose}, which executes \underline{\textbf{F}}eature \underline{\textbf{T}}oken \underline{\textbf{P}}runing for human \underline{\textbf{Pose}} estimation, aiming to push the boundaries of both performance and efficiency in addressing challenging scenes in videos. The proposed method embraces two key innovations: \textbf{(i)} We introduce a multi-scale feature encoder, which comprises a high-resolution branch that primarily focuses on modeling fine-grained spatial correlations within the key frames, and a low-resolution branch that is specifically designed to analyze temporal dynamics across successive frames. \textbf{(ii)} We mathematically formulate a novel density peaks clustering algorithm to perform feature token pruning, which dynamically selects rich visual semantic tokens. This algorithm decouples feature tokens affecting pose results from entangled multi-frame features and reduces computational demands. Empirically, our method significantly and consistently outperforms existing state-of-the-art methods on three large-scale datasets. 

% Comprehensive ablation studies have been conducted, conclusively demonstrating the effectiveness of the designs within our model.

To summarize, the key contributions of this paper are as follows: \textbf{1)} We propose a framework for effective video-based human pose estimation, which encodes multi-grained spatio-temporal dependencies and compensates for detailed semantic diversity from fine-grained features by aggregating multi-grained features. \textbf{2)} 
We propose a novel density peaks clustering algorithm designed to distill key tokens rich in visual cues while pruning away the unimportant ones, which boosts the inference speed and eliminates video redundancy. \textbf{3)} Our approach achieves new state-of-the-art results on three benchmark datasets and provides an interesting insight: discarding redundant features reduces inference cost and simultaneously enhances pose estimation accuracy.

% discarding redundant features accelerates the inference process and simultaneously enhances pose estimation accuracy.

% \begin{figure}[t]
%   \centering
% \includegraphics[width=\linewidth]{case.pdf}
%   \caption{
%   The illustration of recognition accuracy for the same action class under occlusion and normal scenario. \textbf{Left:} Showcase two videos of the same action class, ``Pat on back of other person", under occlusion (top) and normal (bottom) conditions. \textbf{Right:} Bar chart comparing the recognition accuracy of different methods, including MST-GCN, CTR-GCN, FR-Head, HD-GCN, and Ours, in both occlusion and normal scenarios.}
%   %The conceptual diagram depicts joint correlations in Hyperbolic space ({purple}) and Euclidean space ({blue}). The correlation between joint A and joint C ({red dashed curve}) in Hyperbolic space exhibits a non-linear variation compared to that ({blue dashed line}) in Euclidean space.}
% \label{fig1}%%\vspace{-1.5em}
% \end{figure}

\section{Methodology}
\subsection{Problem Formulation}
The method adheres to a top-down paradigm, initially extracting each individual from an image followed by an estimation of their poses. Specifically, the process begins with the deployment of an object detector to extract the bounding box for person $i$ within a video frame $I_{t}$. Thereafter, the bounding box is enlarged by 25$\%$, and the same individual is cropped from the surrounding frames. In this section, we demonstrate the case where the input frame sequence consists of a total of three frames that include the frame $I_{t}^{i}$ to be detected, denoted as $\boldsymbol{\mathcal{I}}_{\boldsymbol{t}}^{\boldsymbol{i}} = \{I_{t-1}^{i}, I_{t}^{i}, I_{t+1}^{i}\}$.

% FTP-Pose develops a multi-scale representation learning framework to aggregate visual details and proposes a feature token pruning technique to enhance efficiency and reduce redundancy.

\textbf{Method overview.} The overview pipeline of the proposed FTP-Pose is illustrated in Figure~\ref{fig:framework}.  First, we utilize a Vision Transformer (ViT) backbone~\cite{dosovitskiy2020imagevit} to extract visual features $\{\boldsymbol{F}_{t-1}^{i}, \boldsymbol{F}_{t}^{i}, \boldsymbol{F}_{t+1}^{i}\}$ from the input sequence $\boldsymbol{\mathcal{I}}_{\boldsymbol{t}}^{\boldsymbol{i}}$, which are then fed into the Multi-Grained Feature Encoder (MGFE). In the MGFE, the high-resolution branch primarily models the detailed spatial correlations of the key frame $I_{t}^{i}$, while the low-resolution branch focuses on the temporal continuity of the input sequence, with both branches performing feature token pruning. Finally, we aggregate the outputs of the two branches and apply a pose head to obtain the final pose heatmaps \(\mathbf{H}_{t}^{i}\). The following sections will detail the key design aspects.

\begin{table*}[t] 
    \centering
    \caption{Comparisons with the state-of-the-art methods for video pose estimation on the validation sets of the PoseTrack2017~\cite{iqbal2017posetrack}, PoseTrack2018~\cite{andriluka2018posetrack}, and PoseTrack2021~\cite{doering2022posetrack21} datasets. Best performances are highlighted in bold.} \label{table:compare}
    \resizebox{0.82\textwidth}{!}{
    \begin{tabular}{c|c|ccccccc|c}
        \hline
        Dataset & Method & Head & Shoulder & Elbow & Wrist & Hip & Knee & Ankle & Mean \\
        \hline
         \multirow{9}{*}{PoseTrack2017 Val Set}   
                             & PoseWarper~\cite{bertasius2019posewarper}  &  81.4 &88.3 &83.9& 78.0& 82.4 &80.5 &73.6 & 81.2  \\
                             &  DCPose~\cite{liu2021dcpose} & 88.0 & 88.7 & 84.1 & 78.4 &  83.0 & 81.4 & 74.2 & 82.8  \\
                             % DetTrack~\cite{wang2020dettrack}& 89.4& 89.7& 85.5 &79.5 &82.4 &80.8 &76.4 &83.8 \\
                            % &   SLT-Pose~\cite{gai2023sltpose} & 88.9& 89.7 &85.6 &79.5 &84.2 &83.1 & 75.8& 84.2\\
                            &   KPM~\cite{fu2023kpm} & 89.5 & 90.0 & 87.6 & 81.8  & 81.1 & 82.6 & 76.1 &84.6\\
                            &   M-HANet~\cite{jin2024mhanet} & 90.3 & 90.7& 85.3& 79.2& 83.4& 82.6& 77.8 &84.8\\
                            &   FAMI-Pose~\cite{liu2022fami} & 89.6  & 90.1 & 86.3 & 80.0 & 84.6 & 83.4 & 77.0 & 84.8   \\
                            &   DSTA~\cite{he2024dsta} & 89.3  & 90.6 & 87.3 & 82.6 & 84.5 & 85.1 & 77.8 & 85.6   \\
                            &   TDMI-ST~\cite{feng2023tdmi}  & \textbf{90.6} & 91.0 & 87.2 & 81.5 &  85.2 & 84.5 & 78.7 & 85.9 \\
                             & DiffPose~\cite{feng2023diffpose} & 89.0 & 91.2 & 87.4 & 83.5 & 85.5 & 87.2 & 80.2 & 86.4  \\
                             &  \textbf{FTP-Pose (Ours)} & 89.5 & \textbf{91.6} & \textbf{88.3} & \textbf{84.7} & \textbf{88.0} & \textbf{87.8} &  \textbf{80.9} & \textbf{87.4} \\

        \hline

        \multirow{9}{*}{PoseTrack2018 Val Set}  & PoseWarper~\cite{bertasius2019posewarper}  & 79.9 & 86.3  &82.4  & 77.5 & 79.8  &78.8 & 73.2  &79.7 \\
                                  & DCPose~\cite{liu2021dcpose}  &84.0& 86.6& 82.7& 78.0& 80.4 &79.3 &  73.8& 80.9 \\
                             % DetTrack~\cite{wang2020dettrack}& 84.9  &87.4  &84.8 & 79.2  &77.6  &79.7  &75.3 & 81.5 \\
                            &  FAMI-Pose~\cite{liu2022fami}  &85.5& 87.7 &84.2& 79.2& 81.4 &81.1 &  74.9& 82.2  \\
                            &  M-HANet~\cite{jin2023hanet}& \textbf{86.7}& 88.9& 84.6& 79.2& 79.7& 81.3& 78.7& 82.7\\
                             & DiffPose~\cite{feng2023diffpose}  &85.0 &87.7& 84.3 &81.5 &81.4& 82.9 &  77.6 &83.0 \\
                            &  KPM~\cite{fu2023kpm} & 85.1& 88.9& 86.4& 80.7& 80.9& 81.5& 77.0& 83.1\\
                            &  DSTA~\cite{he2024dsta} & 85.9  & 88.8 & 85.0 & 81.1 & 81.5 & 83.0 &  77.4 & 83.4   \\
                            &  TDMI-ST~\cite{feng2023tdmi}  &\textbf{86.7}& 88.9 &85.4 &80.6 &82.4& 82.1&  77.6& 83.6  \\ 
                          
                            & \textbf{FTP-Pose (Ours)} & 85.2 & \textbf{89.3} & \textbf{87.1} & \textbf{82.2} & \textbf{84.5} & \textbf{84.3} &  \textbf{79.1} &  \textbf{84.5} \\

        \hline
        \multirow{5}{*}{PoseTrack2021 Val Set} & DCPose~\cite{liu2021dcpose} &83.2 &84.7& 82.3 &78.1 &80.3 &79.2 & 73.5& 80.5  \\
                            & FAMI-Pose~\cite{liu2022fami}  &83.3 &85.4 &82.9 &78.6 &81.3 &80.5 & 75.3 &81.2 \\
                            % & DiffPose~\cite{feng2023diffpose}   &84.7& 85.6 &83.6 &80.8 &81.4& 83.5 &   \textbf{80.0} &82.9 \\
                            & DSTA~\cite{he2024dsta} & \textbf{87.5}  & 87.0 & 84.2 & 81.4 & 82.3 & 82.5 &  77.7 & 83.5   \\
                            & TDMI-ST~\cite{feng2023tdmi}  &86.8 &87.4 &85.1 &81.4 &83.8 &82.7 &  78.0 &83.8 \\
                            & \textbf{FTP-Pose (Ours)} & 86.2 &  \textbf{88.4} & \textbf{85.9} & \textbf{82.5} & \textbf{85.1} & \textbf{83.8} &  \textbf{79.7} &  \textbf{84.5} \\

        \hline
    \end{tabular}
    }
\end{table*}

\subsection{Feature Token Pruning}
A naive approach to capturing temporal dynamics involves concatenating multi-frame feature tokens and processing them through full-token Transformers across the model, which demands considerable computational resources and absorbs substantial task-irrelevant information from redundant tokens. To address this issue, we attempt to perform a feature token pruning operation to discard superfluous tokens, retaining only those with high semantic information. However, selecting representative feature tokens is a challenging task.

% \begin{equation}
% \begin{aligned}
% \rho_{i} = \exp ( -\frac{1}{k} \mathbb{E}_{x_j \sim k\mathrm{NN}(x_i)}  \|x_i - x_j\|_2^2  ),
% \end{aligned}
% \end{equation}

To this end, we introduce a density peaks clustering algorithm~\cite{rodriguez2014clustering}, which is parameter-free, to categorize feature tokens into several groups based on their similarity and to extract the most salient tokens. Token cluster centers are distinguished by their elevated density in comparison to adjacent tokens, coupled with a notably greater separation from other tokens exhibiting higher densities. For a token \( x_i \) within the feature tokens \( x \), the local density \( \rho_i \) can be determined by the following formula: 
\begin{equation}
\begin{aligned}
\rho_{i} = \exp ( -\mathbb{E}_{x_j \sim k\mathrm{NN}(x_i)}  \|x_i - x_j\|_2^2  / \tau ) ,
\end{aligned}
\end{equation}
where \( k \mathrm{NN}(x_{i}) \) represents the neighbor set of \( x_i \), identified through a \( k \)-nearest neighbors algorithm. \(\left\|x_{i}-x_{j}\right\|_{2}^{2}\) denotes the squared Euclidean distance between the tokens \(x_{i}\) and \(x_{j}\). $\tau$ denotes the temperature parameter. 

We then introduce \( \delta_i \) to quantify the shortest distance between token \( x_i \) and other tokens with a higher density. The \( \delta_i \) for the token with the highest density is defined as the greatest distance to any other token. The \( \delta_i \) for each token is computed as follows:
\begin{equation}
\delta_{i} = \min ( \left\{ \left\|x_{i}-x_{j}\right\|_{2} : \rho_{i} < \rho_{j} \right\} \cup \left\{ \max \left\|x_{i}-x_{j}\right\|_{2} \right\} ) .
\end{equation}

% \begin{equation}
% \begin{aligned}
% \delta_{i}=\left\{\begin{array}{l}
% \min _{j: \rho_{i}<\rho_{j}}\left\|x_{i}-x_{j}\right\|_{2}, \text { if } \exists \rho_{i}<\rho_{j} \\
% \max _{j}\left\|x_{i}-x_{j}\right\|_{2}, \text { otherwise } .
% \end{array}\right.
% \end{aligned}
% \end{equation}

We further combine \( \rho \) and \( \delta \) to obtain the clustering center scores. A higher token score indicates greater density and distance, making it more likely to be a cluster center. The score can be calculated as: $score_{i} = \rho_{i} \cdot \delta_{i}$. Given that these cluster centers possess highly salient visual cues, we select representative tokens $\bar{x} \in \mathbb{R}^{(N // \varepsilon ) \times C}$ from the feature tokens $x \in \mathbb{R}^{N  \times C}$ based on the pruning ratio \(\varepsilon\).

\subsection{Multi-Grained Feature Encoder}
Here we propose a Multi-Grained Feature Encoder that includes a high-resolution branch to mine fine-grained spatial details of key frames and a low-resolution branch to capture temporal contexts. Although our framework comprises two branches, they are designed to operate in parallel and share model parameters. Coupled with the adoption of feature token pruning, the speed of our model depends on the low-resolution branch and is faster than without pruning operations.

Given the feature tokens \(\mathbf{F}_{t}^{i}\) of the key frame \(I_{t}^{i}\), we first reshape and upsample it by a factor of four to obtain high-resolution features. Then, we perform token pruning to obtain the pruned tokens and feed them into Transformer blocks to achieve fine-grained feature tokens \(\mathbf{F}_{f}\).

Given a sequence of feature tokens \(\{\mathbf{F}_{t-1}^{i}, \mathbf{F}_{t}^{i}, \mathbf{F}_{t+1}^{i}\}\), we initially extract temporal contexts in a coarse manner through spatio-temporal transformers. We then non-uniformly select representative feature tokens from these temporal features, utilizing feature token pruning. The selected tokens are further refined through Transformer blocks, which share parameters with the high-resolution branch, to produce \(\mathbf{F}_{c}\).

\subsection{Pose Heatmap Generation}
To obtain the final pose heatmaps, we initially merge \(\mathbf{F}_f\) and \(\mathbf{F}_c\) by employing a cross-attention layer:
\begin{equation}
\begin{aligned}
\text{CrossAttn}(\mathbf{F}_{f},\mathbf{F}_{c})  = \sigma ((W_{q}\mathbf{F}_{f} ) (W_{k}\mathbf{F}_{c} )^{T} / \sqrt{d}  )(W_{v}\mathbf{F}_{c} ) ,
\end{aligned}
\end{equation}
where \(\sigma(\cdot)\) denotes the softmax operation, and \(d\) refers to the embedding dimension. The \(W_{q}\), \(W_{k}\), and \(W_{v}\) are all learnable mapping matrices. We then employ a Multi-Layer Perceptron layer (MLP) and a pose head to yield the final pose \(\mathbf{H}_t^i\).

\textbf{Loss function.} Finally, we adopt the standard pose heatmap loss \(\mathcal{L}_{\mathrm{H}}\) to supervise the final predicted pose heatmaps $\mathbf{H}_{t}^{i}$ to converge to the ground-truth pose heatmaps $\mathbf{G}_{t}^{i}$:
\begin{equation}\label{loss}
\begin{aligned}
 \mathcal{L}_{\mathrm{H}} = \left\| \mathbf{H}_{t}^{i} - \mathbf{G}_{t}^{i}\right\|_{2}^{2}.
\end{aligned}
\end{equation}

\section{Experiments}

\subsection{Datasets and Implementation Details}
We perform extensive experiments to evaluate the efficacy of our method across three large-scale datasets: PoseTrack2017~\cite{iqbal2017posetrack}, PoseTrack2018~\cite{andriluka2018posetrack}, and PoseTrack2021~\cite{doering2022posetrack21}. The input image size is fixed at 256\(\times\)192 and the patch size is set to 16. Our model is trained on a single RTX 4090 GPU for a total of 20 epochs, starting with an initial learning rate of \(2 \times 10^{-3}\), which is decreased by a factor of ten at the 16th epoch. All data augmentation techniques are similar to those used in TDMI~\cite{feng2023tdmi}. To assess the performance of the model in human pose estimation, we compute the average precision (AP) for each joint and subsequently aggregate these values to derive the mean average precision (mAP).

\subsection{Comparison with State-of-the-Art Methods}
We evaluate the performance of the proposed FTP-Pose against the latest state-of-the-art (SOTA) methods, including M-HANet~\cite{jin2023hanet}, TDMI-ST~\cite{feng2023tdmi}, DSTA~\cite{he2024dsta}, and others, on three larger-scale datasets: PoseTrack2017~\cite{iqbal2017posetrack}, PoseTrack2018~\cite{andriluka2018posetrack}, and PoseTrack2021~\cite{doering2022posetrack21}. All experimental results on the three validation datasets are presented in Table~\ref{table:compare}.

\begin{figure}[t]
\centering
\includegraphics[width=.90\linewidth]{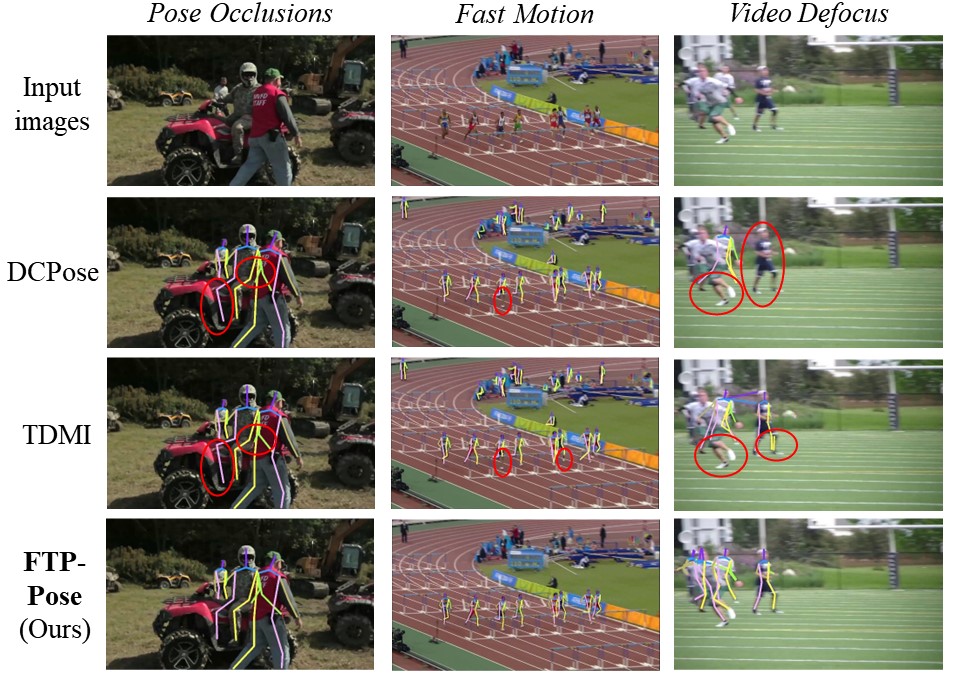}
\caption{Qualitative comparison of our FTP-Pose, DCPose~\cite{liu2021dcpose}, and TDMI~\cite{feng2023tdmi} on the PoseTrack dataset, featuring challenges such as pose occlusions, fast motion, and video defocus. Red solid circles denote the inaccurate pose results.} 
\label{fig:comparison}
\end{figure}

\textbf{PoseTrack2017.} 
We first benchmark our method on the PoseTrack2017 dataset, and the results demonstrate that our model consistently surpasses existing methods, achieving an mAP of 87.4 ($\uparrow$1.0). Especially, we achieve promising improvements for the more challenging joints, with an mAP of 84.7 ($\uparrow$ 1.2) for wrists and an mAP of 80.9 ($\uparrow$ 0.7) for ankles. \textbf{PoseTrack2018.} FTP-Pose also surpasses the latest state-of-the-art method TDMI-ST~\cite{feng2023tdmi}, attaining an mAP of 84.5 ($\uparrow$ 0.9), with an mAP of 82.2 ($\uparrow$ 1.6), 84.3 ($\uparrow$ 2.2), and 79.1 ($\uparrow$ 1.5) for the wrist, knee, and ankle, respectively. For \textbf{PoseTrack2021}, our method has improved by 0.7 mAP compared to the current best TDMI-ST~\cite{feng2023tdmi}. Encouragingly, for the difficult joints such as wrist and ankle, we have still achieved improvements of 1.1 and 1.7, respectively. In addition, we present visualized results comparing our method with existing methods for challenging scenarios in Figure~\ref{fig:comparison}, demonstrating our method's robustness.

\begin{table}[t] 
    \centering
    \fontsize{9pt}{9pt}\selectfont 
    \renewcommand{\arraystretch}{1.2} % 设置行间距
    \caption{Ablation of different
designs in the FTP-Pose framework.}
    \resizebox{0.82\linewidth}{!}{
    \begin{tabular}{c|ccc}
        \hline
        \textbf{Method}  & \textbf{Param.} & \textbf{mAP}  &  \textbf{Speed (FPS)}  \\
        \hline
        Baseline& 72M & 85.9 & 16   \\
        \hline
        % + multi-scale framework  & 86.8 & {1.4 ($\uparrow$)} \\
        % + feature token pruning & 87.4& {2.0 ($\uparrow$}) \\
        (a) & 78M & 86.7 ($\uparrow$0.8) &    14 ($\downarrow$12.5\%)  \\
        (b) & 78M & \textbf{87.4} ($\uparrow$\textbf{1.5}) &   \textbf{31}  ($\uparrow$\textbf{93.8\%})  \\
        %\hline
        \hline
    \end{tabular}}
    \label{table:component}
\end{table}

\subsection{Ablation Study}

In this section, we perform ablation studies to evaluate the impact of individual design components and various pruning ratios on the PoseTrack2017 validation dataset. 

%we perform ablation experiments to examine the influence of each component in our method. We also explore the effects of modifying masking and noising ratios. These ablation studies are conducted on the cross-subject benchmark of NTU-RGB+D 120 using the joint input.

\textbf{Study on components of FTP-Pose.} We experimentally evaluate the contribution of each component in the proposed method, as shown in Table~\ref{table:component}. We first establish a baseline by connecting a low-resolution branch without feature token pruning to the backbone. (a) We apply the multi-grained learning framework to the baseline. Although the number of parameters slightly increases and the speed slightly decreases, the mAP is improved by 0.8. (b) We apply feature token pruning to the model, once again pushing the performance boundary to 87.4 mAP. Additionally, we observe that the inference speed is nearly doubled ($\uparrow$ \textbf{93.8}\%) compared to the baseline, demonstrating that our token pruning operation can not only eliminate redundant information but also greatly accelerate the inference process.

\textbf{Study on different pruning ratios.} We further explore the impact of different pruning ratios \(\varepsilon\) for the high- and low-resolution branches. As depicted in Table~\ref{ratio}, our method achieves optimal performance when the high-resolution branch (HRB) ratio and the low-resolution branch (LRB) ratio are both set to 6. This value is empirically determined through comprehensive experimental validation, demonstrating that it offers the best trade-off between computational efficiency and accuracy in pose estimation. Lower ratios tend to preserve excessive redundant information, hindering performance, whereas higher ratios risk discarding valuable tokens along with the noise. This analysis underscores the importance of carefully selecting pruning ratios to maximize the benefits of token pruning while preserving critical information.

%We analyze that lower pruning ratios retain a significant amount of redundant information, thereby affecting the estimation performance, while higher ratios discard too many tokens, leading to the inadvertent removal of tokens carrying useful information.

\begin{table}[t] 
\centering
\fontsize{9pt}{10pt}\selectfont 
\caption{The experimental mAP results across various pruning ratios.}
\resizebox{0.70\linewidth}{!}{
\begin{tabular}{cccccc}
\hline
\multirow{2}{*}{\textbf{Ratios}} & & \multicolumn{4}{c}{\textbf{HRB }} \\
\cline{3-6}
& & {1} & {3} & {6} & {10} \\
\hline
\multirow{4}{*}{\textbf{LRB }} & {1} & 86.7 & 86.9 & 86.5 & 85.8  \\
&{3} & 86.1 &  86.2 &  86.9 & 86.7  \\
 &{6} & 85.3 & 86.5 & \textbf{87.4} & 87.2  \\
 &{10} & 85.5 & 86.2 & 87.1 & 86.6 \\
\hline
\end{tabular}}\label{ratio}
\end{table}

%compenonts, ratiomotion-reconstruction task and the motion-denoising task
% \subsection{Robustness Analysis in Challenging Scenarios (RQ3)}
% To evaluate the robustness of our method in challenging scenarios, we randomly \emph{mask} or \emph{add noise to} the input motion sequence during the testing phase and compare its performance against SoTA methods on the cross-subject benchmark of NTU-RGB+D 120. The results in Fig.~\ref{fig:quantitative} reveal that: \textbf{(1)} Our method achieves the best performance across all mask and noise ratios. \textbf{(2)} As the ratios increase, the performance gap between our method and SoTA methods widens, indicating that our method is more robust in challenging scenarios.
% \begin{figure}[t]
% %\begin{minipage}[t]{0.51\linewidth}
% \centering
% \includegraphics[width=.98\linewidth]{quantitative.pdf}
% \caption{Comparison of model performance in challenging scenarios.}\label{fig:quantitative}%\vspace{-1em}
% \end{figure}

\section{Conclusion}

This research introduces a novel approach for video-based human pose estimation, delivering remarkable improvements in both accuracy and computational efficiency. By developing an innovative multi-scale learning framework integrated with a feature pruning strategy, we address the complex challenges inherent in capturing fine-grained spatial contexts and high-dimensional temporal dynamics. A major contribution of this work is the introduction of the density peaks clustering algorithm, which intelligently selects representative feature tokens. By pruning redundant feature tokens, we have significantly expedited the inference process without sacrificing accuracy. Extensive experiments across three benchmark datasets confirm the superiority of our method, demonstrating marked improvements in both accuracy and efficiency. This research offers a fresh perspective to the field, showing that discarding redundant features can accelerate computation while simultaneously improving pose estimation accuracy. 
% These insights highlight the potential for more efficient, scalable solutions in video-based pose estimation, making it a valuable contribution to ongoing advancements in the field.

%In this paper, we present an efficient multi-scale learning framework that integrates a feature pruning strategy for video-based human pose estimation. Specifically, the framework simultaneously explores fine-grained spatial contexts and captures high-dimensional temporal dynamics. We further introduce a density peaks clustering algorithm to select representative feature tokens and perform token pruning to reduce redundant information. Extensive experiments demonstrate that our method achieves significant improvements in both performance and efficiency across three benchmark datasets.
\bibliographystyle{IEEEtran}
\bibliography{ICASSP}
\end{document}